\pdfoutput=1

\documentclass[11pt]{article}

\usepackage[preprint]{acl}
\usepackage{times}
\usepackage{latexsym}

\usepackage[T1]{fontenc}

\usepackage[utf8]{inputenc}

\usepackage{microtype}

\usepackage{inconsolata}

\usepackage{graphicx}

\usepackage{booktabs}
\usepackage{enumitem}

\title{Evaluating Contextualized Representations of (Spanish) Ambiguous Words: A New Lexical Resource and Empirical Analysis}

\author{Pamela D. Rivière$^{1}$, Anne L. Beatty-Martínez$^{1}$, Sean Trott$^{1,2}$ \\
         $^{1}$Department of Cognitive Science, UC San Diego\\
         $^{2}$Computational Social Science, UC San Diego\\
         \texttt{\{pdrivier, abeattymartinez, sttrott\}@ucsd.edu}}

\begin{document}
\maketitle
\begin{abstract}

Lexical ambiguity---where a single wordform takes on distinct, context-dependent meanings--serves as a useful tool to compare across different language models' (LMs') ability to form distinct, contextualized representations of the same stimulus. Few studies have systematically compared LMs' contextualized word embeddings for languages beyond English. Here, we evaluate semantic representations of Spanish ambiguous nouns in context in a suite of Spanish-language monolingual and multilingual BERT-based models. We develop a novel dataset of minimal-pair sentences evoking the same or different sense for a target ambiguous noun. In a pre-registered study, we collect contextualized human relatedness judgments for each sentence pair. We find that various BERT-based LMs' contextualized semantic representations capture some variance in human judgments but fall short of the human benchmark. In exploratory work, we find that performance scales with model size. We also identify \textit{stereotyped trajectories} of target noun disambiguation as a proportion of traversal through a given LM family's architecture, which we partially replicate in English. We contribute (1) a dataset of controlled, Spanish sentence stimuli with human relatedness norms, and (2) to our evolving understanding of the impact that LM specification (architectures, training protocols) exerts on contextualized embeddings.

\end{abstract}

\section{Introduction}\label{sec:intro}
State-of-the-art language models (LMs) display a remarkable level of formal linguistic competence \citep{mahowalddissociating}. To date, however, we currently lack a precise accounting of the mechanisms underlying LMs' fundamental linguistic capabilities. The opacity of model internals has motivated work probing the transformations that inputs undergo as they are processed through various model components \citep{tenney-etal-2019-bert, hu2020systematic, wang2022interpretability, zou2023representation}. This work has focused on LMs trained with English-language corpora, with a smaller subset of studies investigating cross-linguistic representations in multilingual models \citep{chang-etal-2022-geometry,wendler2024llamas, michaelov-etal-2023-structural} or comparing representations across multiple monolingual models \citep{edmiston2020systematic}. As others have noted \cite{blasi2022over, bender-2009-linguistically}, an over-reliance on English as a ``model language'' limits the generalizability of findings, as well as potential applications \cite{blasi-etal-2022-systematic}. Here, we extend interpretability work to Spanish, a language spoken by almost 600M people (with almost 500M native speakers)\footnote{\url{https://www.exteriores.gob.es/en/PoliticaExterior/Paginas/ElEspanolEnElMundo.aspx}}. Specifically, we: (1) evaluate the trajectory of ambiguous Spanish words' semantic representations within mono- and multilingual LMs and (2) identify layers along a model's architecture whose semantic representations best capture human judgments of semantic relatedness. 

Lexical ambiguity---where a given wordform evokes multiple related or unrelated meanings---offers a unique opportunity to dissociate a word's form from the contextualized, semantic representations that it can take on as it interacts with a given model's architecture. Specifically, we can evaluate whether and how LMs integrate the surrounding lexical items in a sentence to produce flexible, context-dependent representations. The representation and processing of ambiguous words is also well-studied in humans \cite{rodd2004modelling, martin1999strength, duffy1988lexical}, offering a convenient comparison group. Finally, ambiguity appears to pervade language, with some estimates in English positing that more than 80\% of words have multiple meanings \citep{rodd2004modelling}; more frequent words are also more likely to evoke multiple senses \citep{zipf1945meaning,piantadosi2012communicative}. Ambiguity, then, is an important \textit{phenomenon} to contend with in LMs and a useful \textit{tool} to understand LM representations.

We first provide a survey of related work on ambiguity and interpretability in LMs (Section \ref{sec:related}), then present a novel dataset of human relatedness judgments about Spanish ambiguous words---in context (SAW-C) (Section \ref{sec:dataset}). Section \ref{sec:primary_results} documents our use of this dataset to empirically probe the representation of ambiguous words in pre-trained spanish LMs, focusing first on pre-registered analyses\footnote{Code and data to run all analyses are available at: \url{https://github.com/seantrott/spanish_norms}. Preregistration is available at: \url{https://osf.io/n5htp}.} of the monolingual BETO \citep{CaneteCFP2020}. We next systematically compare multiple monolingual Spanish and multilingual BERT-based models (Section \ref{sec:multiple_models}). We conclude with a discussion of the implications of the current results (Section \ref{sec:discussion}), as well as limitations and directions for future work (Section \ref{sec:limitations}).


       

\section{Related work}\label{sec:related}

To date, lexical ambiguity has been largely explored within English monolingual models \citep{haber-poesio-2020-word,trott-bergen-2021-raw,delong2023offline}. Less work is available with models trained on other languages, such as Spanish \citep{garcia-2021-exploring,gari-soler-apidianaki-2021-lets}, and crucial distinctions emerge across studies within this literature: (1) the operationalization of \textsc{same} and \textsc{different sense} conditions, (2) the degree to which sentential context is controlled around the target word, and (3) the extent to which human semantic judgments are collected to set usage-based expectations for the \textit{context-dependence} and \textit{graded quality} of ambiguous word meanings \citep{erk-etal-2013-measuring, trott2023word}. 

In Spanish, mono- and multilingual BERT-based models can capture information about semantic relationships between homonyms and their synonyms \citep{garcia-2021-exploring} and can approximate words' degree of polysemy \citep{gari-soler-apidianaki-2021-lets}. However, available studies tend to leverage naturalistic sentence stimuli from sense-annotated corpora. While valuable, the variability in token sequence length and target word position within naturalistic sentential contexts may make it challenging to isolate the precise effect of the context's semantic \textit{content} from the uncontrolled effects of sentence frame \citep{haber-poesio-2020-word}. We thus follow as closely as possible the experimental structure leveraged (for English) in \citet{trott-bergen-2021-raw}, creating Spanish-language sentence pairs that vary along a single context cue evoking either the same or different sense of the target ambiguous noun (see Section \ref{sec:materials}). For our dataset, we document the extent to which context cues presented true minimal pairs (e.g. zero-token-differences across sentence pairs) for the BERT-based models we tested (see \textbf{Appendix \ref{sec:appendix_llm-specs-and-tokendiffs}; Table \ref{tab:lm_token_diffs}}).

Using more controlled sentence stimulus design, coupled with empirically collected human benchmarks, prior studies in English have shown that BERT-based models' contextualized embeddings capture some---though not all---variance in human similarity \citep{haber-poesio-2020-word} and relatedness \citep{trott-bergen-2021-raw,delong2023offline} judgments for ambiguous English words. Some of this work has also argued that the \textit{continuous} nature of LM contextualized representations makes them well-suited as models of human word meanings, which are likely graded to some extent \cite{elman2009meaning, li2024probing, li2021word, trott2023word, rodd2004modelling, nair-etal-2020-contextualized,d-zamora-reina-etal-2022-black}---though importantly, may also exhibit marked \textit{categoriality} compared to LM representational spaces \cite{trott2023word}.







Finally, another important line of work has used techniques like classifier probes \cite{tenney-etal-2019-bert} and activation patching \cite{NEURIPS2022_6f1d43d5, wang2022interpretability, merullo2024circuit} to decode the putative functional role of different model components (e.g., layers, attention heads, etc.) in producing observed behavior. 

To our knowledge, there is little to no work combining these strands of research: i.e., making use of graded human judgments about ambiguous Spanish words to trace the dynamics of contextualized representations in pre-trained Spanish language models. This is the gap we aim to address.

\section{Human Annotation Study}\label{sec:dataset}

We created a dataset containing graded human judgments about Spanish ambiguous words---in context (SAW-C). This process involved first producing and curating materials (i.e., target words and sentences), collecting judgments from native Spanish speakers, and validating those judgments for quality and reliability. This study was pre-registered on the Open Science Framework (OSF) platform.

\subsection{Materials}\label{sec:materials}

All sentences were developed by two native (Puerto Rican) Spanish speakers, both authors in this study. $102$ target words were drawn from noun lists collected in earlier studies of Spanish lexical ambiguity \citep{monzo1991estudio,fraga2017saw}, as well as spontaneously generated and then verified using the online Real Academia Espa\~nola\footnote{\url{https://www.rae.es/}} dictionary. We excluded wordform meanings that corresponded to distinct parts of speech, but we accepted a small fraction ($8/102$) of nouns whose grammatical gender changed across meanings. Each target ambiguous noun was embedded within sentence pairs that differed by a single modifier\footnote{For $187/812$ sentence pairs, the modifiers varied along more than just a single word, as was the case when the modifiers required different prepositions, or the grammatical gender of the modifier differed across sentences and required distinct determiners and contractions. Of these, $173$ pairs differed by $1$ word; $14$ differed by $2$ words.}, termed context cue\footnote{Context cues were adjectival modifiers for $100/102$ target nouns; verbs for $2/102$.}. The average number of words in sentences was $4.72$. The context cues across the sentence pair could evoke either the \textsc{Same} or \textsc{Different} sense of the word across the contexts. 

\begin{description}[noitemsep]
    \item \textbf{1a.} Compró el \textbf{aceite} de oliva (\textit{[S/he] bought the olive oil})
    \item \textbf{1b.} Compró el \textbf{aceite} de cocina (\textit{[S/he] bought the cooking oil})
    \item \textbf{2a.} Compró el \textbf{aceite} de motor (\textit{[S/he] bought the motor oil})
    \item \textbf{2b.} Compró el \textbf{aceite} de carro (\textit{[S/he] bought the car oil})
\end{description}

The minimum number of \textit{sentence pairs} per target noun was $6$, with a maximum of $28$ (M = $7.96$). A total of $812$ sentence pairs and $102$ target nouns were included in the dataset.

Finally, for the purpose of human norming, the $812$ sentence pairs were assigned to $10$ experimental lists using a Latin Square design, where each list had approximately $81$ or $82$ sentence pairs.

\subsection{Participants}\label{subsec:ppts}

Our goal was to collect a minimum of $10$ judgments per sentence pair (i.e., a minimum of $100$ participants). Because we anticipated a non-zero exclusion rate, our pre-registration specified: 1) an initial goal of $120$ participants; and 2) a plan to sample more participants as needed, if any sentence pairs had fewer than $10$ observations after applying the pre-registered exclusion criteria.

Using the two-step process described above, we recruited an initial pool of $139$ participants through Prolific. Participants received $\$2.40$ for participating and the median completion time was 12 minutes and 23 seconds, for an average rate of $\$11.64$ per hour. On Prolific, we screened for participants who reported that their primary language was Spanish; we specifically recruited participants from the United States, as well as countries in which the dominant language was Spanish (including Chile, México, and Spain).

We excluded participants (1) who failed ``catch'' trials (where the sentences in the pair were identical), (2) whose task completion times exceeded the sample mean by 3 standard deviations, (3) whose inter-annotator agreement for the items they rated was very low (Pearson's $r^2 < 0.1$), and (4) who self-identified as non-native Spanish speakers. After all exclusions, we considered data from a total of $131$ participants.

Participants' self-identified nationalities corresponded heavily to México (69), Spain (39), and Chile (20); 1 participant was from the United States and 2 participants were from Venezuela.\footnote{As described in Section \ref{sec:validation}, we found relatively high correlations between mean relatedness judgments produced by participants across Chile, México, and Spain.} 54 participants self-identified as female (74 male, 3 non-binary). The average age was $30.97$ (median = $28$), and ranged from $20$ to $59$.

\subsection{Procedure}

After providing consent, participants were given instructions explaining that some words can have different meanings in different contexts (using an example that was \textit{not} included in the experimental materials), and that the goal of the experiment was to collect ratings about the relatedness of the meanings expressed by a given word across two sentence contexts.

\begin{figure}
    \centering
    \includegraphics[width=0.75\linewidth]{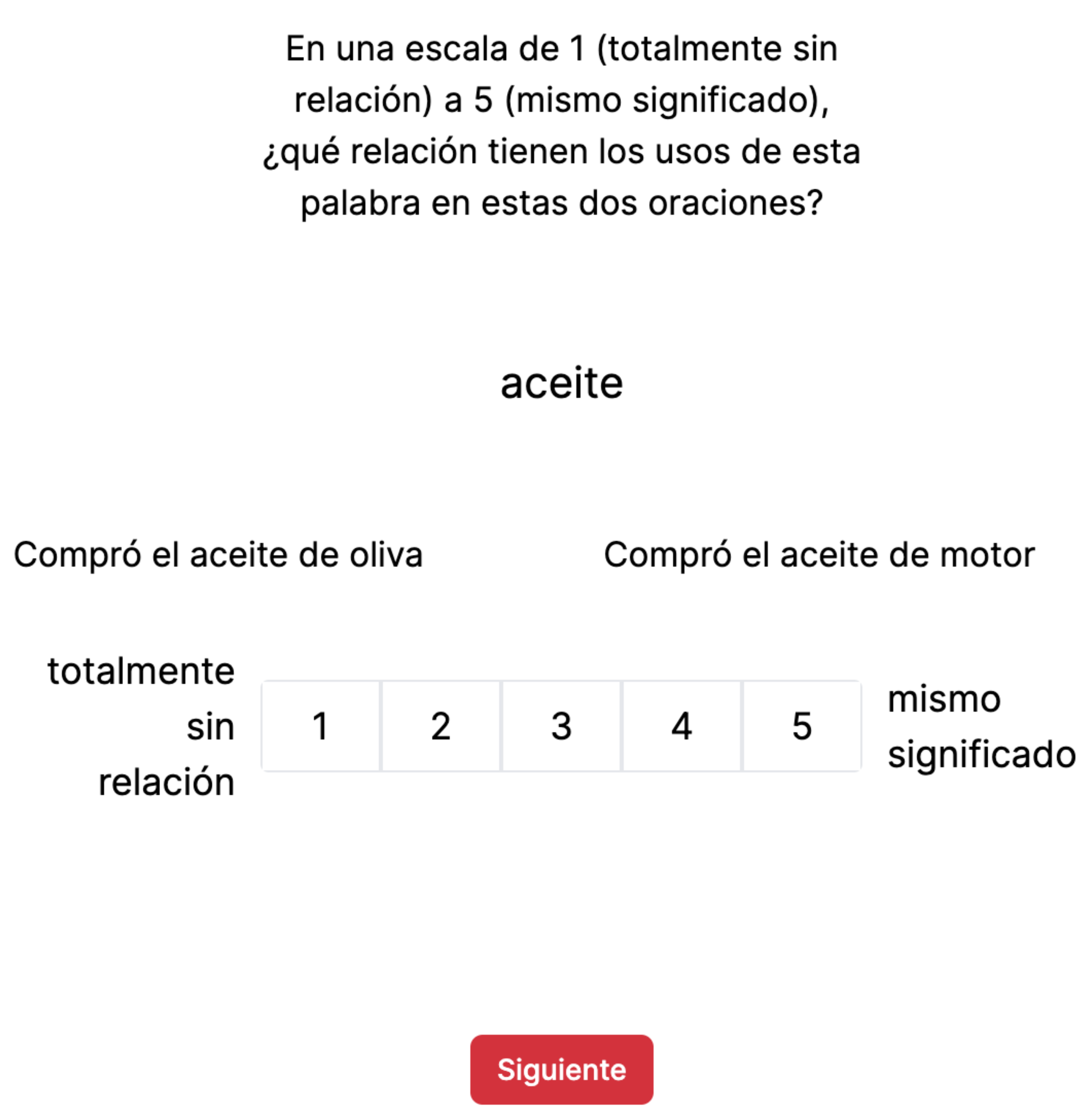}
    \caption{Sample item from task. Participants emitted graded relatedness judgments from a scale of 1 (totally unrelated) to 5 (same sense) for a given target word (here: \textbf{aceite} -- \textit{oil}), using information provided by the context cue across the sentence pair (here: \textbf{de oliva} / \textbf{de motor} -- \textit{olive} / \textit{motor}).}\label{fig:sawc_task}
\end{figure}

Each sentence pair was presented on a separate page. Participants were instructed to rate each word on a scale from 1 (\textit{totalmente sin relación}, ``totally unrelated'') to 5 (\textit{mismo significado}, ``same meaning'')\footnote{For alternative ordinal annotation schemes in relatedness datasets, see \cite{schlechtweg2018diachronic,schlechtweg-etal-2024-durel,schlechtweg2025comedi}}. The target word (e.g., \textit{aceite}, ``oil'') was centered in larger font, and the target sentences were presented side-by-side (the side was randomized across trials). Participants indicated their response via button-press; see \textbf{Figure \ref{fig:sawc_task}} for an example. We also included one ``catch'' trial in the experiment, which simply contained the same sentence, repeated (i.e., the correct answer was $5$).

The entire experiment (including consent form, instructions, and debrief page) was conducted in Spanish. Participants were randomly assigned to lists, and the order of items within each list was randomized. After completing the primary experiments, participants read a debrief form explaining once more that the goal of the experiment was to collect judgments about ambiguous Spanish words, and that their data would be anonymized before analysis and publication.

\subsection{Validation of Final Dataset}\label{sec:validation}

We validated the final dataset using several approaches. First, we applied multiple exclusion criteria (Section \ref{subsec:ppts}) and collected a minimum of $10$ ratings per sentence pair. The average number of ratings per pair was $13.1$; the maximum was $17$.

After applying exclusion criteria, we recalculated inter-annotator agreement to estimate the reliability of the ratings in the final dataset. Following past work \citep{hill2015simlex, trott-bergen-2021-raw} we calculated inter-annotator agreement using a \textit{leave-one-annotator-out} scheme. For each of the final $131$ participants, we calculated Spearman's $\rho$ between the judgments produced by that participant and the mean judgment for those same sentence pairs, leaving out that participant's data. The resulting distribution of correlation coefficients ranged from $0.39$ to $0.88$, with an average correlation of $0.77$. This number is comparable to past work using similar methods \cite{hill2015simlex, trott-bergen-2021-raw}.

Finally, we compared average relatedness judgments across the three main demographic groups reported by participants (Chile, México, and Spain). Judgments across each group were all strongly correlated ($r > 0.82$ in all cases).

\subsection{Relatedness of Same vs. Different Sense Contexts}\label{sec:rq1}

We then asked whether and how human relatedness judgments varied as a function of whether two uses of a word (e.g., ``aceite'') corresponded to the \textsc{Same Sense} or \textsc{Different Sense}. These will hereafter be considered the levels of the binary variable we call Sense Relationship.

\begin{figure}
    \centering
    \includegraphics[width=\linewidth]{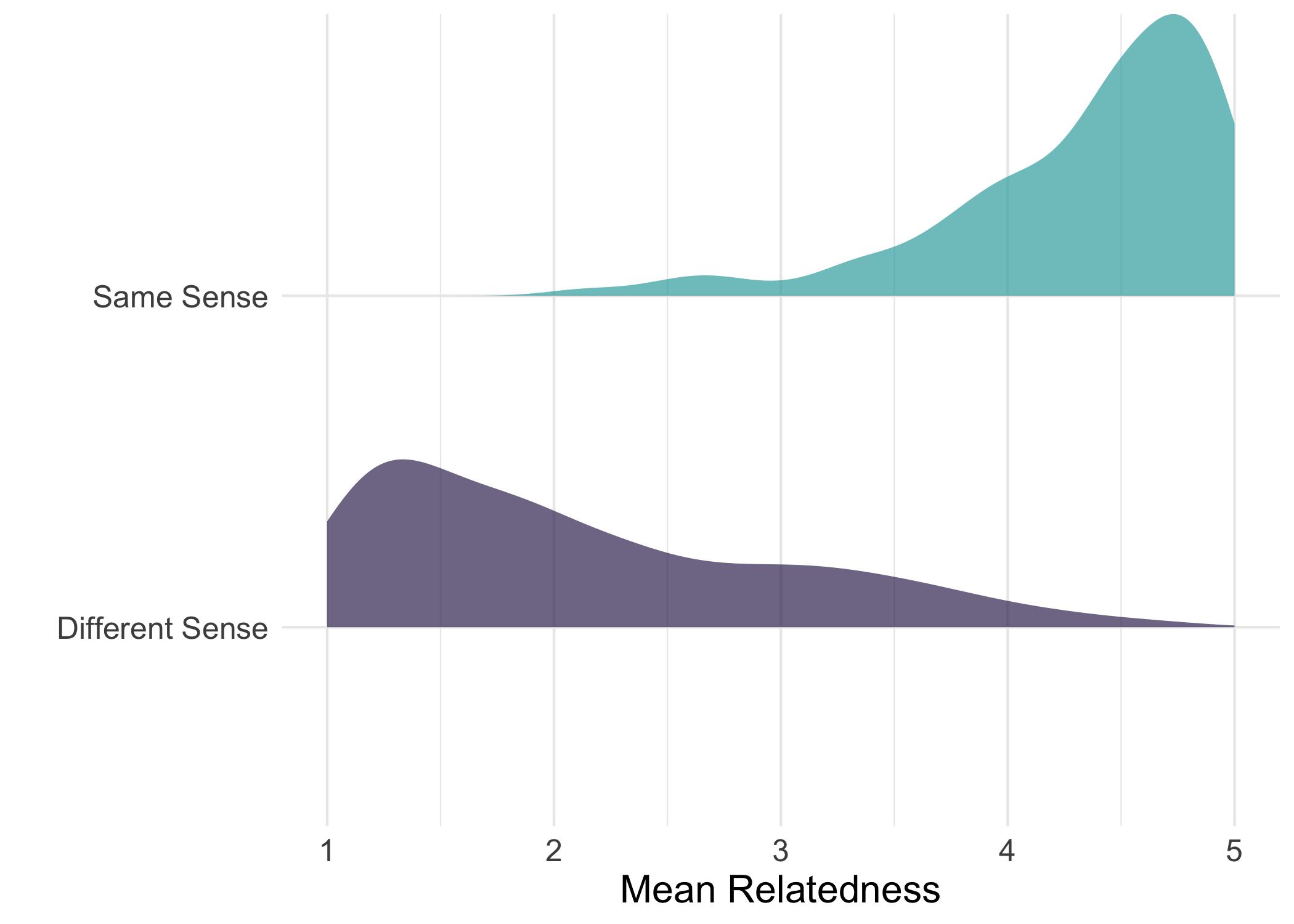}
    \caption{Density plot representing the distribution of mean relatedness judgments for sentence pairs. As expected, word meanings were rated as more related when used in \textsc{Same} Sense than \textsc{Different} Sense contexts.}
    \label{fig:rq1_density}
\end{figure}

Using the entire dataset of trial-level judgments ($10639$ observations), we fit a linear mixed effects model in R using the \textit{lme4} package \cite{de2011estimation}, which had Relatedness as a dependent variable and a fixed effect of Sense Relationship. The model also contained by-participant and by-list random slopes for the effect of \textsc{Same Sense}, and random intercepts for participants, lists, and words. (The specification of random effects was determined by beginning with the maximal model, then reducing as needed for model convergence \cite{barr2013random}.) The full model explained significantly more variance than a reduced model omitting only the effect of Sense Relationship  $[\chi^2(1) = 39.59, p < .001]$. As expected, \textsc{Same} Sense contexts were rated as more related on average ($M = 4.35, SD = 1.14$) than \textsc{Different} Sense contexts ($M = 2.11, SD = 1.41$; \textbf{Figure \ref{fig:rq1_density})}.

\section{Analysis of BETO, a Pre-Trained Spanish LLM}\label{sec:primary_results}

Using SAW-C as a probe, we conducted multiple pre-registered analyses on how BETO---a pre-trained monolingual Spanish BERT-based model \cite{CaneteCFP2020}---represents ambiguous words in context, whether and to what extent these representations are predictive of human semantic representations, and \textit{which layers} of the model contained the most information \cite{tenney-etal-2019-bert}. \textbf{Table \ref{tab:rqs}} summarizes the research questions and their results.

\subsection{Model Details}

Our pre-registered analyses used the cased version of a Spanish monolingual BERT-based model: BETO, comprised of 12 layers, each made up of 12 self-attention heads, and hidden size of 768, trained on a corpus of approximately 3B words \citep{CaneteCFP2020}\footnote{Accessed via \url{https://huggingface.co/dccuchile/bert-base-spanish-wwm-cased}}. Exploratory analyses leveraged the models summarized in the \textbf{Appendix \ref{sec:appendix_llm-specs-and-tokendiffs}} (\textbf{Table \ref{tab:lm_comparison}}). Each sentence in the dataset (bracketed by special tokens [CLS] and [SEP]) was tokenized separately according to each model's tokenizer. Sentences contained periods, unlike the sentences viewed by human participants\footnote{In a previous version of this study, we presented models with sentences that did not contain periods, to match the stimuli exactly as human participants saw them. However, we found that LMs' outputs explained more variance in human judgments when the periods were included at the end of the sentences.}. We report sentence pair tokenization differences across models in \textbf{Table \ref{tab:lm_token_diffs}} (\textbf{Appendix \ref{sec:appendix_llm-specs-and-tokendiffs}}). When the target noun was represented by multiple subword tokens, we took the average embedding across tokens. We extracted embeddings from each model layer.

\subsection{Which layer of BETO best captures sense boundaries?}\label{sec:rq6}

We first assessed which layers of BETO produced representations that best distinguished between \textsc{Same Sense} and \textsc{Different Sense} uses of word. To address this question, we calculated the Cosine Distance between these contextualized representations of the target word from each sentence pair for each layer. Concretely, this yielded $812$ Cosine Distance values for each layer of BETO.

\begin{figure}
    \centering
    \includegraphics[width=\linewidth]{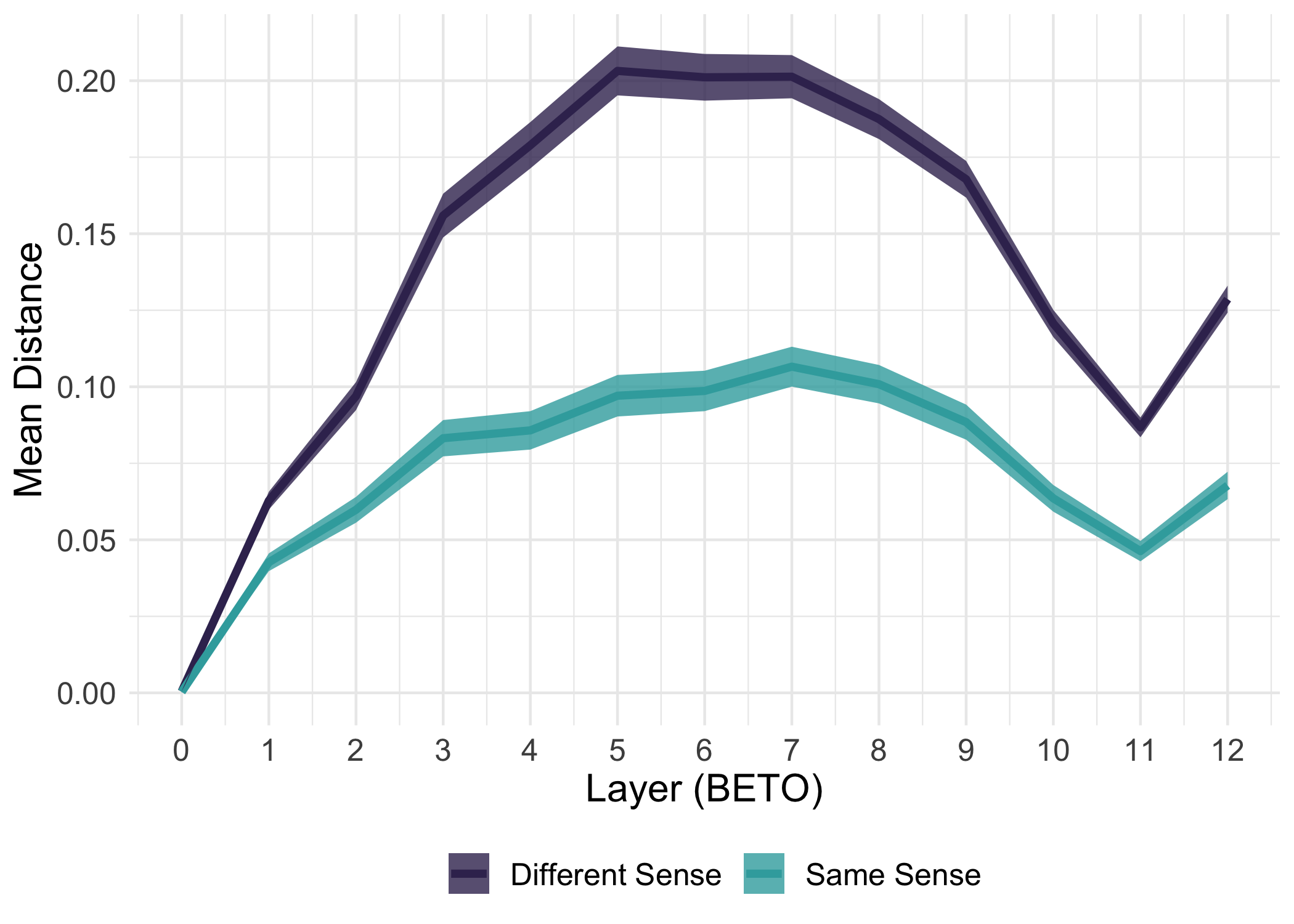}
    \caption{Average Cosine Distance between the contextualized representations of the target ambiguous word across each layer of BETO, depicted as a function of whether the contexts cued the \textsc{Same Sense} or \textsc{Different Sense}.}
    \label{fig:rq6_beto_same_different}
\end{figure}

Then, we asked how Cosine Distance evolved through the network's layers with respect to the \textsc{Same/Different Sense} distinction. We built a series of logistic regression models in R with Sense Relationship as a dependent variable, and Cosine Distance from a given layer $\ell_i$ of BETO as an independent variable. We then measured the Akaike Information Criterion (or $AIC$) \cite{akaike2011akaike, burnham2004multimodel} of the resulting model as a measure of model fit. The best-fitting model used Cosine Distance from \textit{layer 6}. \textbf{Figure \ref{fig:rq6_beto_same_different}} highlights the change in Cosine Distance across layers of BETO as a function of Sense Relationship, suggesting that the \textit{difference} between conditions was largest at this layer.

\subsection{Which layer of BETO best predicts relatedness?}\label{sec:rq2}

Our second question was whether certain layers of BETO produced representations that better predicted \textit{human relatedness judgments} than others. For each layer, we calculated the correlation coefficient (both Pearson's \textit{r} and Spearman's \textit{$\rho$}) between Cosine Distance values obtained from BETO and the distribution of Mean Relatedness judgments obtained for each sentence pair. We also calculated $R^2$ as an estimate of the amount of \textit{variance explained} in human relatedness judgments as a function of Cosine Distance from that layer alone.

As depicted in \textbf{Figure \ref{fig:rq2_r2_by_layer}}, the layer of BETO with the highest $R^2$ was \textit{layer 12} ($R^2 = 0.36, r = -.60, \rho = -.61$). However, performance did not meaningfully improve beyond \textit{layer 7} ($R^2 = 0.35$). This suggests that the operations performed by later layers were less useful in terms of producing contextualized representations that captured relevant variance in relatedness judgments (see also \textbf{Appendix \ref{sec:appendix_expected_layer}}).

\subsection{BETO under-performs inter-annotator agreement}\label{sec:rq4}

\begin{figure}
    \centering
    \includegraphics[width=\linewidth]{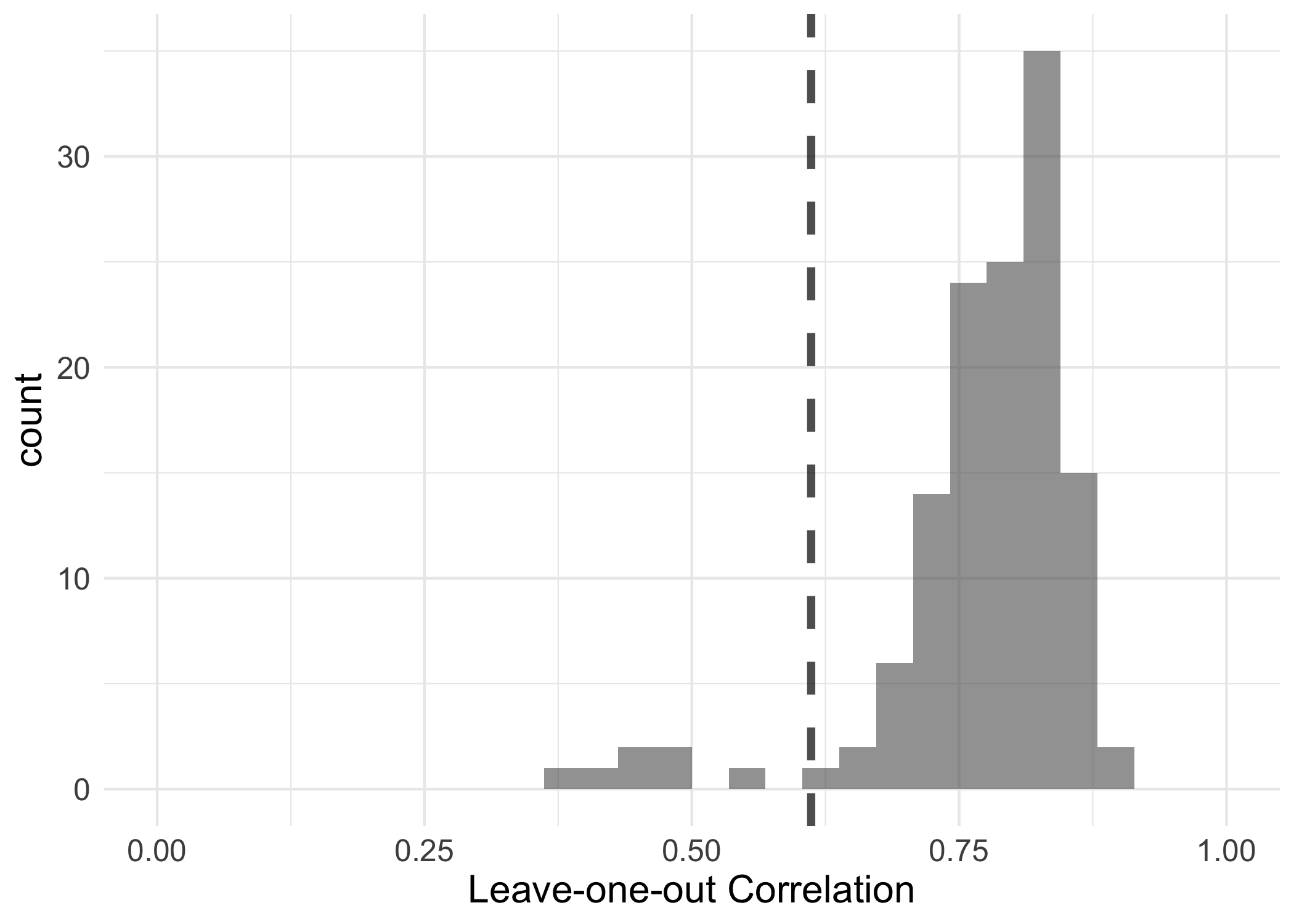}
    \caption{Distribution of human inter-annotator agreement scores, calculated using a \textit{leave-one-annotator out} scheme. The vertical dashed line represents the correlation between human judgments and Cosine Distance values extracted from BETO.}
    \label{fig:agreement_with_beto}
\end{figure}

We then compared the best-performing layer ($\ell = 12$) to human inter-annotator agreement (calculated in Section \ref{sec:validation}). The correlation magnitude of the best-performing layer ($\rho = 0.61$) was considerably lower than the average inter-annotator agreement ($\bar{X}_\rho = 0.77$), with BETO's performance lying approximately in the bottom $5\%$ of the distribution of agreement values (\textbf{Figure \ref{fig:agreement_with_beto}}).

\subsection{BETO is Less Sensitive to Sense Boundaries}\label{sec:rq3}

Past work conducted in English \cite{trott-bergen-2021-raw, trott2023word} suggests that LMs are less sensitive to sense boundaries---the distinction between \textsc{Same} and \textsc{Different Sense}---than humans. However, it is unclear whether this effect generalizes to Spanish speakers and Spanish LMs.  

We constructed a linear mixed effects model in R with Relatedness as a dependent variable, fixed effects of Cosine Distance and Sense Relationship, by-participant random slopes for both fixed effects, and random intercepts for participants, words, and lists. The full model explained significantly more variance than a model omitting only the effect of Sense Relationship  $[\chi^2(1) = 325.21, p < .001]$. The full model also explained more variance than a model omitting only Cosine Distance $[\chi^2(1) = 236.73, p < .001]$. Further, the $R^2$ of a linear regression model predicting Mean Relatedness using Sense Relationship alone ($R^2 = .61$) explained more variance than Cosine Distance alone ($R^2 = 0.36$); adding both predictors resulted in a modest improvement over the Sense Relationship model ($R^2 = 0.66$).

\begin{figure}
    \centering
    \includegraphics[width=\linewidth]{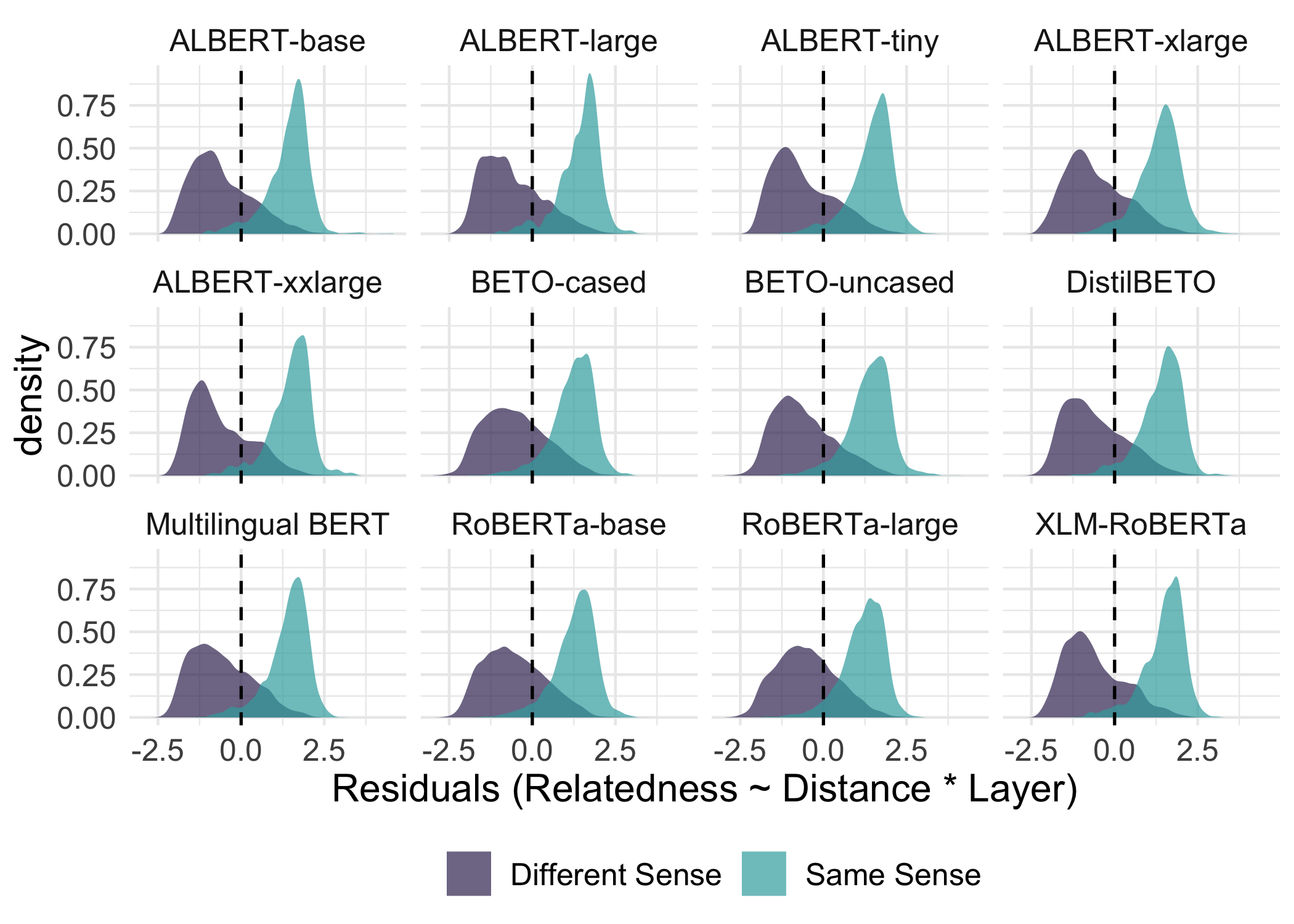}
    \caption{Residuals of linear regression models fit for each LM, predicting relatedness from the interaction between cosine distance and layer position; residual distributions are separable as a function of Sense Relationship.}
    \label{fig:sense_boundary_residuals}
\end{figure}

Finally, we extracted the \textit{residuals} of a linear regression model predicting Mean Relatedness from Cosine Distance alone. We then plotted the distribution of these residuals according to Sense Relationship. As illustrated in \textbf{Figure \ref{fig:sense_boundary_residuals}}, BETO (as well as all the Spanish language models tested in Section \ref{sec:multiple_models}) consistently \textit{underestimated} the relatedness of \textsc{Same} sense pairs, and consistently \textit{overestimated} the relatedness of \textsc{Different} sense pairs.

\begin{table}[ht]
\centering
\begin{tabular}{@{}p{0.5\columnwidth}cc@{}}
\toprule
\textbf{Research Question} & \textbf{Result} & \textbf{Section} \\
\midrule
Do humans judge \textsc{Same Sense} uses to be more related than \textsc{Different Sense} uses? & Yes & \ref{sec:rq1} \\
Which layer of BETO is most sensitive to the \textsc{Same/Different Sense} distinction? & $\ell = 6$ & \ref{sec:rq6} \\
Which layer of BETO is most correlated with human relatedness judgments? & $\ell = 12, 7$ & \ref{sec:rq2} \\
Does BETO match human inter-annotator agreement?  & No & \ref{sec:rq4} \\
Does BETO ``explain away'' the effect of categorical Sense Relationship in humans (e.g. ``sense boundaries'')? & No & \ref{sec:rq3} \\
\bottomrule
\end{tabular}
\caption{Summary of pre-registered research questions and their results.}
\label{tab:rqs}
\end{table}

\section{Comparing Pre-Trained Spanish Language Models}\label{sec:multiple_models}

Section \ref{sec:primary_results} tested several pre-registered hypotheses (\textbf{Table \ref{tab:rqs}}) with respect to a single pre-trained Spanish LM. Here, we extend this work in exploratory analyses of additional pre-trained Spanish BERT-based models. Testing multiple models is an important step towards establishing the \textit{external validity} of a finding; additionally, it is useful for testing hypotheses about model scale \cite{kaplan2020scaling} or other model specifications (e.g., architecture, multilingual status).

\subsection{Models}

We considered $10$ monolingual Spanish language models (including BETO) and $2$ multilingual models. Models varied in their training procedures (e.g., BERT vs. RoBERTa; \citet{liu2019roberta}), tokenization scheme, number of layers, training corpus size, total number of parameters, and whether or not they were multilingual (\textbf{Table \ref{tab:lm_comparison}}).

\subsection{Impact of model scale}

Past work \cite{kaplan2020scaling} suggests that increases in a model's number of parameters may correlate with metrics of model performance (e.g., perplexity). At the same time, there is some evidence that increasing scale does not always produce more human-like representations, i.e., large models with lower perplexity do not always better predict human reading times \cite{kuribayashi2021lower}. 

\begin{figure}
    \centering
    \includegraphics[width=\linewidth]{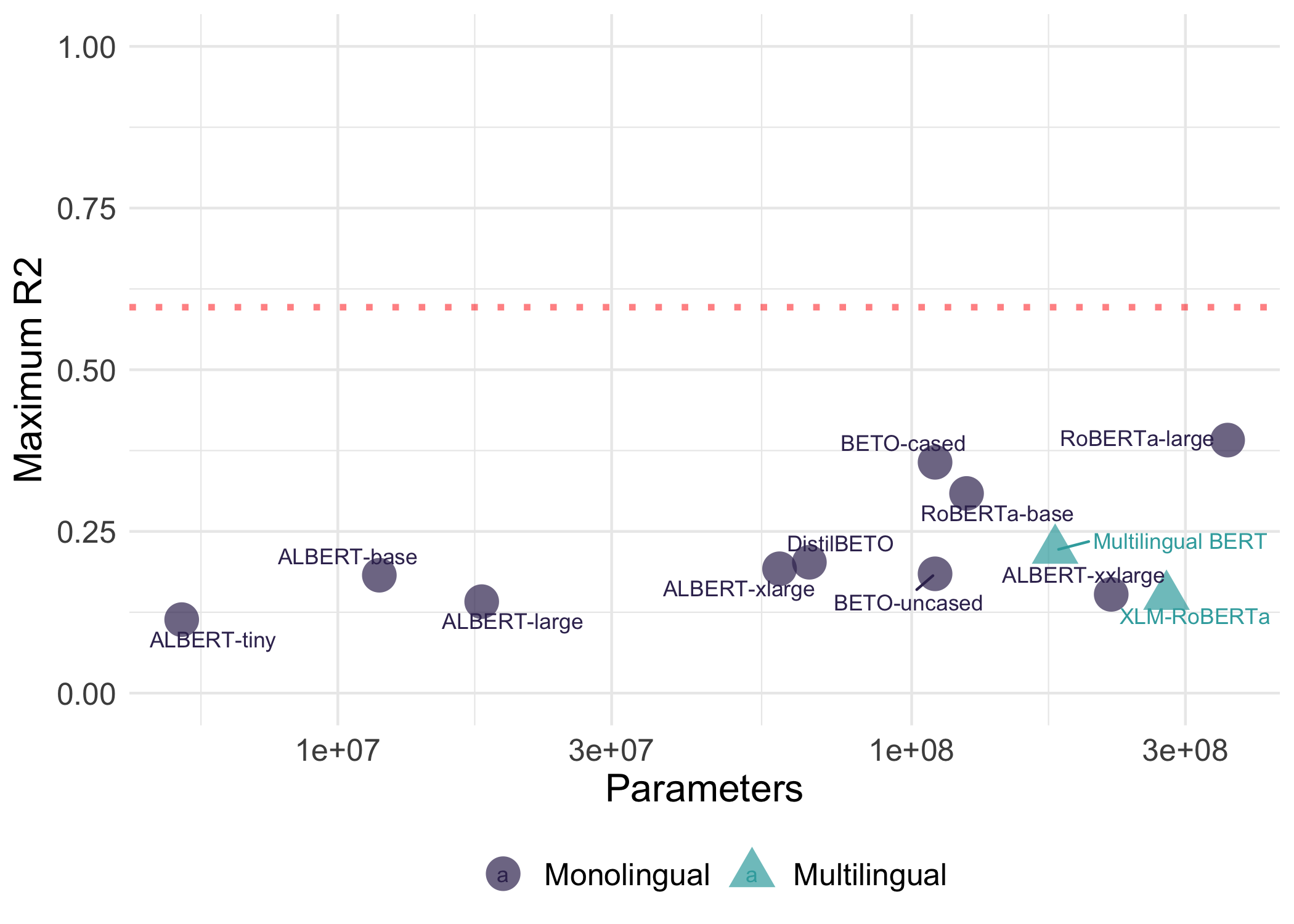}
    \caption{Maximum $R^2$ achieved by each model by number of parameters and multilingual status. Horizontal dashed line depicts the average variance explained in a leave-one-annotator-out scheme.}
    \label{fig:r2_by_params}
\end{figure}

To assess this question, we compared the maximum $R^2$ achieved by each of the $12$ language models\footnote{I.e., the $R^2$ from the best-performing layer of each model.} and asked whether a model's best $R^2$ was related to its size. We found that model size was in fact correlated with its ability to predict human relatedness judgments in Spanish \footnote{Though, we note that this becomes true only when models are tested with stimuli containing periods at the end of the sentences. The RoBERTa model family was particularly sensitive to this implementational difference.} (\textbf{Figure \ref{fig:r2_by_params}}). RoBERTa-large (with the most parameters) outperformed all other models. Multilingual models, however, performed just as poorly as monolingual models with many fewer parameters. Notably, models within the Spanish-language ALBERT family---carefully controlled for various training and architectural properties \cite{canete-etal-2022-albeto}---failed to show clear, consistent evidence of scaling beyond the jump from -tiny to -base architectures. None of the models matched the variance explained in average judgments by individual human annotators (i.e., inter-annotator $R^2$).

\subsection{Performance across layers}

In Section \ref{sec:primary_results}, we found that cosine distance measures extracted from the \textit{middle and last layers} of BETO were most useful for predicting whether two contexts belonged to the same sense, and for predicting human relatedness judgments. Do other models or model families show the same trajectory of performance across layers?

\begin{figure}
    \centering
    \includegraphics[width=\linewidth]{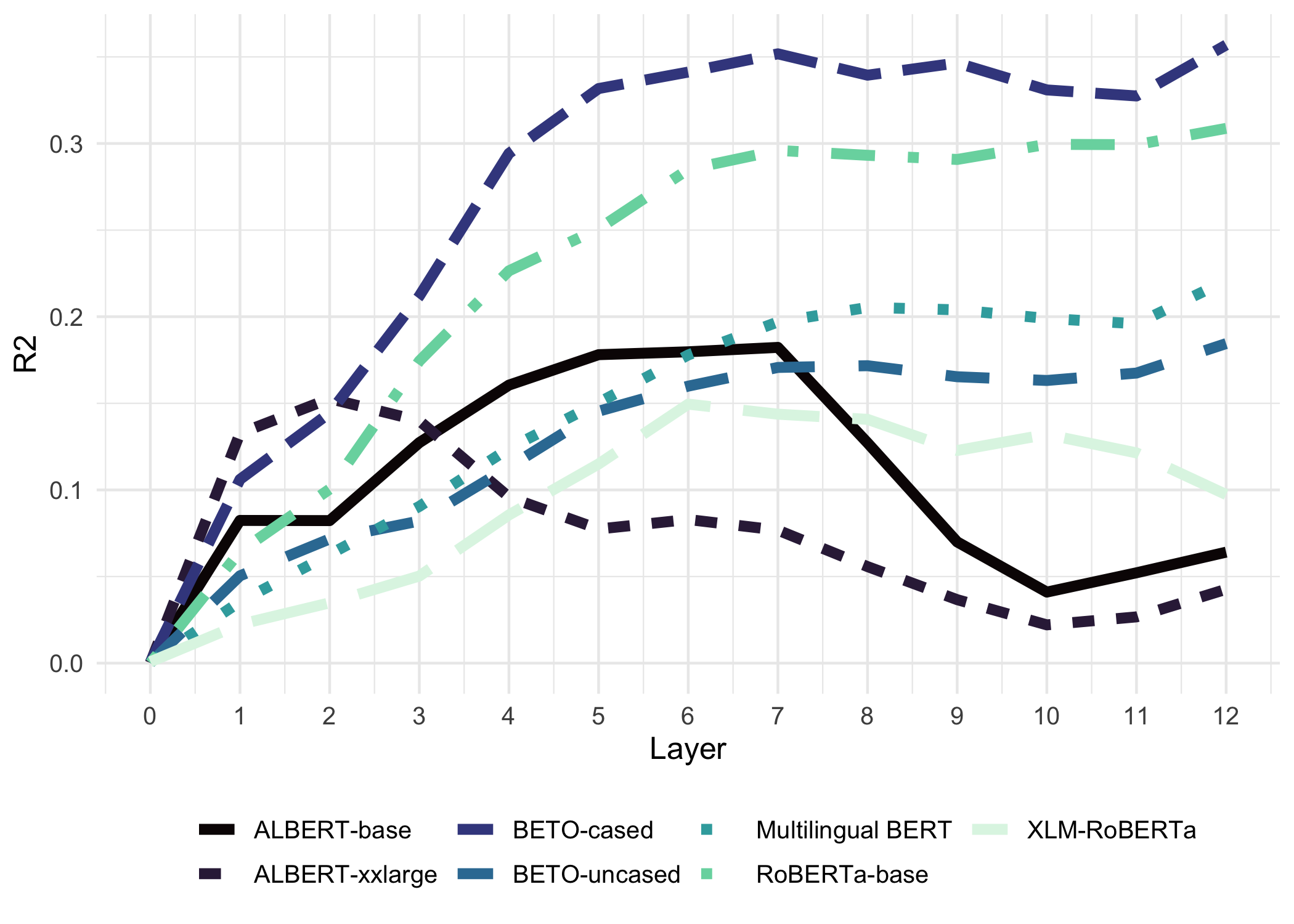}
    \caption{Depiction of seven pre-trained Spanish LMs' ability to predict human relatedness judgments across layers (measured as $R^2$). For ease of illustration, this plot shows only LMs with $12$ layers.}
    \label{fig:rq2_r2_by_layer}
\end{figure}

\begin{figure}
    \centering
    \includegraphics[width=\linewidth]{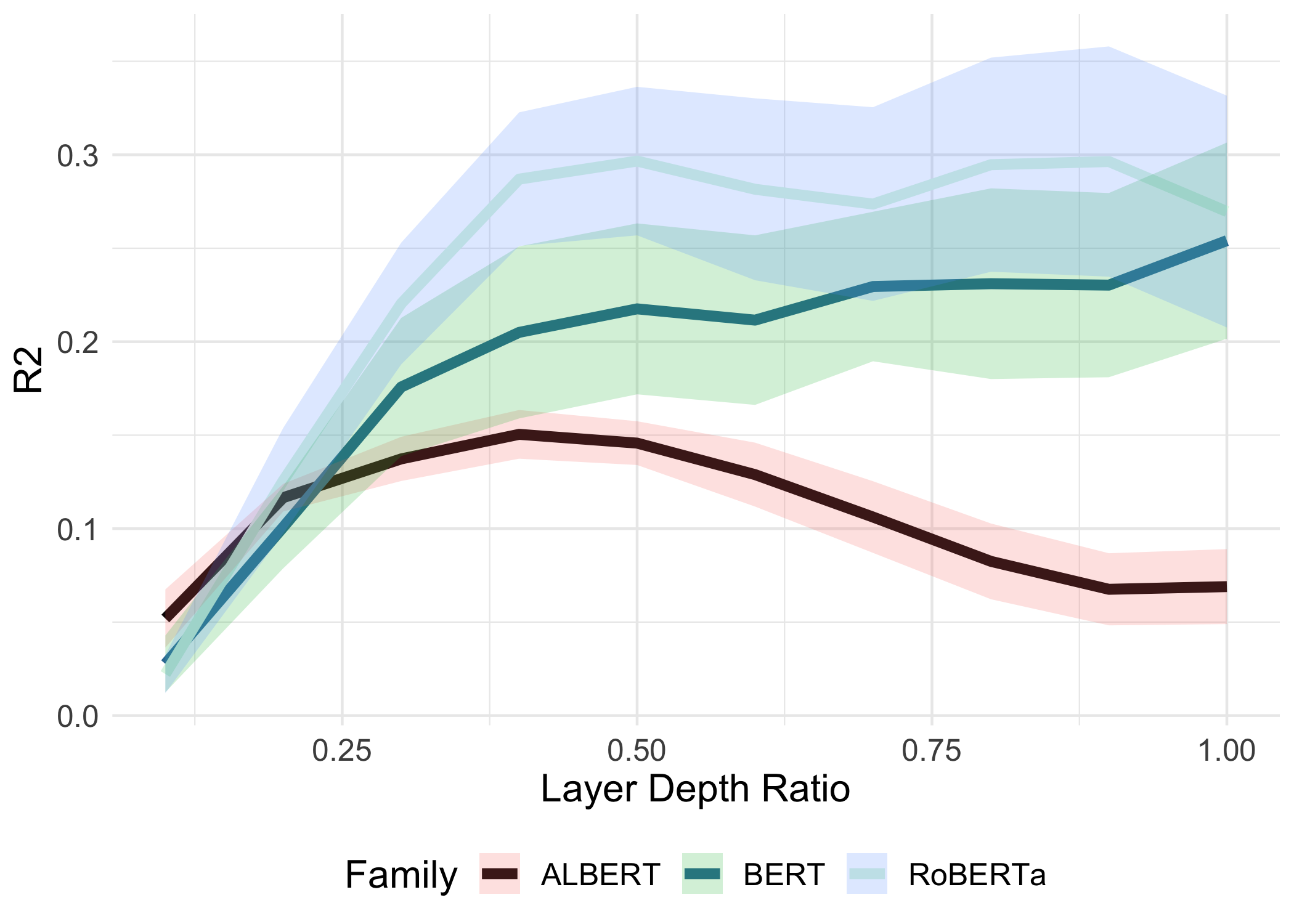}
    \caption{Depiction of LM ability to predict mean relatedness judgments (measured as $R^2$), broken down by \textit{Model Family} and \textit{Layer Depth Ratio}, i.e., with each layer divided by the the total number of layers in a given model. 4 BERT models (1 multilingual), 3 RoBERTa models (1 multilingual), and 5 ALBERT models.}
    \label{fig:r2_by_prop_layer}
\end{figure}

Models varied in their number of layers. Thus, we first compared the trajectory of $R^2$ across layers for the subset of models with the same number of layers, i.e., $12$ layers (\textbf{Figure \ref{fig:rq2_r2_by_layer}}) or $24$ layers (\textbf{Figure \ref{fig_appepndix:r2_by_layer}}). In each case, we identified two qualitatively distinct ``classes'' of trajectory: a \textit{rise and plateau} trajectory, in which performance improves up until a point (e.g., layer 6) and then stays relatively stable; and a \textit{rise and fall} trajectory, in which performance improves and then decays substantively in the final layers (\textbf{Figure \ref{fig:r2_by_prop_layer}}).

In order to compare all models on the same axis, we calculated the \textit{layer depth ratio}, which divides each layer position in a given network by the total number of layers in that network. We then visualized the average $R^2$ by \textit{layer depth ratio} across the three model families tested: ALBERT, BERT, and RoBERTa. As \textbf{Figure \ref{fig:r2_by_prop_layer}} suggests, the two putative trajectories appear to covary with model family: the ALBERT family of models shows a \textit{rise and fall} trajectory, while the BERT and RoBERTa family of models shows a \textit{rise and plateau} trajectory.

\subsection{Sensitivity to sense categoriality across models}

Lastly, we asked whether the LMs tested \textit{explained away} the sense boundary effect in humans. Each model showed the same pattern as BETO: Cosine Distance \textit{underestimated} the relatedness of \textsc{Same Sense} meanings and \textit{overestimated} the relatedness of \textsc{Different Sense} meanings (\textbf{Figure \ref{fig:sense_boundary_residuals}}).

\section{Discussion}\label{sec:discussion}

We introduced a novel dataset (SAW-C) containing human relatedness judgments about ambiguous Spanish words in controlled, minimal pair contexts. Using this dataset, we probed pre-trained BERT-based models' representations of ambiguous words, finding that: 1) LM representations correlate with human judgments but do not match inter-annotator agreement, and exhibit \textit{systematic errors}; 2) performance varies across layers, with model families showing distinct \textit{trajectories} of performance; and 3) performance scaled with model \textit{size}. 

The systematic \textit{underestimation} and \textit{overestimation} errors observed with respect to \textsc{same} vs. \textsc{different} sense contexts (\textbf{Figure \ref{fig:sense_boundary_residuals}}) is consistent with past work conducted in English \citep{trott-bergen-2021-raw, trott2023word}. One explanation for this is that the initial (static) embedding for an ambiguous wordform might entangle all of its multiple meanings \cite{grindrod2024transformers}, which must then be ``teased apart'' in context---but which might nevertheless persist as ``attractors'' in subsequent layers. Disambiguation could be made even more difficult by the presence of minimal pair contexts \cite{garcia-2021-exploring}. 

Psycholinguistic research suggests that humans also activate uncued, dominant meanings in certain tasks \cite{duffy1988lexical, martin1999strength}; here, however, humans appeared to distinguish target meanings with relative ease. It is possible that humans represent distinct homonymous (though not necessarily polysemous) meanings along clearly differentiable regions of meaning space \citep{rodd2004modelling,trott2023word,trott-bergen-2021-raw,haber-poesio-2020-word}, which would be consistent with the fact that categorical sense boundaries (\textsc{Same} vs. \textsc{Different sense} conditions) explained an overwhelming share of human relatedness judgments (Section \ref{sec:rq3}, \citet{trott2023word}). 

Lastly, this work contributes to expanding the linguistic diversity of both human-annotated benchmarks and interpretability research. Although Spanish ranks among the most widely spoken languages, it suffers from a surprising dearth of resources (with notable exceptions, \citet{baldissin-etal-2022-diawug}), pre-trained models, and interpretability research---particularly when compared to English. In one study, a sample of 550 corpora (spanning 22 languages) contained  \textgreater $50\%$ English-language corpora, while \textless $10\%$ represented Spanish-language corpora \cite{anand2020widening}. A wider research perspective---considering varied languages---is critical for ensuring the generalizability of findings.

\section{Limitations}\label{sec:limitations}

\subsection{Limitations of the Dataset}

SAW-C is limited in \textit{scope}, containing only $812$ sentence pairs. This is considerably smaller than English benchmarks such as BLiMP \cite{warstadt2020blimp}, which contain tens of thousands of examples. However, it is larger than or comparable in size to other, more targeted datasets involving crowd-sourced human annotations \cite{erk-etal-2013-measuring,haber-poesio-2020-word,trott-bergen-2021-raw}. Relative to specifically Spanish-language datasets, ours is the only one we are aware of that collects human judgments for target ambiguous words embedded \textit{within} minimal pair stimuli\footnote{For examples of other Spanish-language datasets that collect human judgments under different experimental conditions, see  \citet{monzo1991estudio,gomez2010estudio,dominguez2001100,haro2017semantic,fraga2017saw,d-zamora-reina-etal-2022-black,baldissin-etal-2022-diawug}.}. Importantly, SAW-C includes not only the sentence pairs but also over $10000$ validated \textit{human judgments} from $131$ participants about those sentence pairs.

Another limitation concerns stimulus generation. The ambiguous nouns included in our dataset were spontaneously produced by native (Caribbean) Spanish speakers, or selected from previously published lists \citep{monzo1991estudio,fraga2017saw}, rather than via automated searches. Spontaneous production of ambiguous words may (1) overrepresent homonymous words \citep{monzo1991estudio}, and (2) underrepresent words whose multiple meanings have large dominance asymmetries \cite{duffy1988lexical}. In future work, we intend to collect dominance norms for these items. Differences in Spanish varieties may have also led some items to appear awkward to annotators from different nationalities \citep{baldissin-etal-2022-diawug,lipski2014many}. We address this concern by showing strong correlations in relatedness judgments from the three heavily represented groups (Chile, Spain, and M\'exico, Section \ref{sec:validation}), but other groups' relatedness judgments may not be represented in this sample. 

Finally, our sentence pairs are not \textit{naturalistic}. Our controlled minimal pair design was intentional, enabling us to identify key differences between LM and human representations, e.g., LMs display less sensitivity than humans to the manipulation of word meaning across minimal pair contexts (\textbf{Figure \ref{fig:sense_boundary_residuals}}). At the same time, it is important to know whether and to what extent the current results replicate with naturalistic stimuli. Thus, we aim to augment SAW-C with naturalistic examples of the target ambiguous words.

\subsection{Limitations of the Analysis}

All the analyses presented here are essentially \textit{correlational}. As others have noted \cite{hewitt-liang-2019-designing, niu2022does, zhou2021directprobe}, supervised methods for probing LM representations are more informative about the ability of the \textit{probe} to learn specific features than the question of whether the LM ``naturally'' encodes that feature and deploys it for token prediction. Future work would benefit from the selective application of ``knock-out'' methods or ``activation patching'' \cite{NEURIPS2022_6f1d43d5}, both of which have proven more successful in characterizing the causal, mechanistic role of model components. We view the current work as a useful starting point, which can motivate future work isolating the mechanistic role of specific model circuits within each layer. 

The finding that distinct model families exhibit distinct \textit{trajectories} in performance ( \textbf{Figure \ref{fig:r2_by_prop_layer}}) is intriguing, but due to its exploratory nature, it is unclear to what extent this finding is reliable and robust to different datasets or probing methods. In a supplementary analysis (see \textbf{Appendix \ref{fig_appepndix:r2_trajectories_english}}), we found qualitatively similar clusters of trajectories in pre-trained English models, though these differences appeared considerably weaker than in the Spanish models. Future work could build on the question of whether---and more importantly, \textit{why}--- distinct model \textit{architectures} and training schemes lead to different \textit{processing mechanisms}.

We also note that analysis of performance as a function of model size was not an ideal test of the scaling hypothesis, given that many features of these models (e.g., training data, model architecture) were not controlled, except within the Spanish ALBERT family. 

Lastly, the structure of Spanish modifiers (which \textit{follow} the target noun) meant that we could only test encoder (e.g. bidirectional) models in this study. To offset this limitation, we tested a large suite of Spanish-trained encoder models, but future studies may leverage stimuli modified to suit decoder-only architectures. 

\section{Ethical Considerations}

This research was conducted with IRB approval. All data from human participants has been fully anonymized before analysis and publication.

\section*{Acknowledgments}
Pamela D. Rivi\`ere is supported by UCSD's Chancellor's Postdoctoral Fellowship Program. We thank Benjamin Bergen and Cameron Jones for valuable discussions.


\bibliography{anthology,anthology2,custom}

\appendix

\section{Appendix}
\label{sec:appendix}

\subsection{LLM Specifications and Dataset Tokenization}\label{sec:appendix_llm-specs-and-tokendiffs}

See \textbf{Table \ref{tab:lm_comparison}} for a summary of the LMs considered in Section \ref{sec:multiple_models}, including architecture, multilingual status, corpus size, tokenization scheme, training objective, number of layers, and number of parameters. Because models used different tokenizers, we also calculated summary statistics about the number of tokens in each sentence in each sentence pair, as well as the average number of \textit{token differences} (i.e., 5 vs. 4 tokens across the members of a given pair) for each tokenizer (see \textbf{Table \ref{tab:lm_token_diffs}}).

\begin{table*}[htpb]
\centering
\begin{tabular}{@{}lccllcc@{}}
\toprule
\textbf{Model} & \textbf{\# Lang} & \textbf{Corpus} & \textbf{Tok.} & \textbf{Trn Obj} & \textbf{\# Layers} & \textbf{\# Params} \\
\midrule
BETO & 1 & $\sim$ 3B & SP & DM, WWM & 12 & $\sim$ 110M \\
BETO-uncased & 1 & $\sim$ 3B & SP & DM, WWM & 12 & $\sim$ 110M \\
mBERT & 104 & ? & WP & MLM, WWM$^{?}$, NSP & 12 & $\sim$ 178M \\
\midrule 
DistilBETO & 1 & $\sim$ 3B & SP & DistilLoss, MLM & 6 & $\sim$ 66M \\
\midrule
ALBETO-tiny & 1 & $\sim$ 3B & SP & MLM & 4 & $\sim$ 5M \\
ALBETO-base & 1 & $\sim$ 3B & SP & MLM & 12 & $\sim$ 12M \\
ALBETO-large & 1 & $\sim$ 3B & SP & MLM & 24 & $\sim$ 18M \\
ALBETO-xlarge & 1 & $\sim$ 3B & SP & MLM & 24 & $\sim$ 59M \\
ALBETO-xxlarge & 1 & $\sim$ 3B & SP & MLM & 12 & $\sim$ 223M \\
\midrule
RoBERTa-BNE-base & 1 & $\sim$ 135B & byte-BPE & MLM (DM$^{?}$) & 12 & $\sim$ 125M \\
RoBERTa-BNE-large & 1 & $\sim$ 135B & byte-BPE & MLM (DM$^{?}$) & 24 & $\sim$ 355M \\
XLM-RoBERTa & 100 & ? & SP & MLM & 12 & $\sim$ 278M \\

\bottomrule
\end{tabular}
\caption{Spanish language model properties and training procedures. Models are cased (distinguish between upper and lowercase characters) unless otherwise specified. All monolingual models are trained on Spanish-language corpora; multilingual models include Spanish-language corpora. \textbf{\textit{Model Notes:}} For mBERT, (a) the corpus size per language varied and we are unsure of the total corpus size$^{13}$, (b) it was unclear to us whether the current version of the model on HuggingFace is updated with the whole-word masking (WWM) technique$^{14}$ during pre-training. For RoBERTa-BNE models, it was unclear to us whether authors used dynamic masked (DM) modeling, as in the English RoBERTa. For XLM-RoBERTa, the corpus size per language may have varied, but we were uncertain to what extent pretraining text was sampled proportionally to its representation in the corpus \citep{conneau-etal-2020-unsupervised}. \textbf{\textit{Acronyms:}} \textbf{BNE:} Biblioteca Nacional de Espa\~na (National Library of Spain); \textbf{byte-BPE:} byte-level Byte-Pair Encoding; \textbf{DistilLoss:} Distillation loss \citep{sanh2019distilbert,canete-etal-2022-albeto}; \textbf{DM:} Dynamic Masking \citep{liu2020roberta}; \textbf{MLM:} Masked Language Modeling; \textbf{NSP:} Next Sentence Prediction; \textbf{SP:} SentencePiece; \textbf{WWM:} Whole-Word Masking; \textbf{WP:} WordPiece. This summary represents our best attempt at gathering and reconstructing some model specifications---they are necessarily incomplete, and may contain inaccuracies borne from either a lack of knowledge regarding more recent updates or an imperfect understanding of the training protocols as described in the relevant primary literature and repositories.} 
\label{tab:lm_comparison}

\end{table*}

\footnotetext[13]{See \url{https://github.com/google-research/bert/blob/master/multilingual.md} for available details on language-specific corpus size selection.}
\footnotetext[14]{See \url{https://github.com/google-research/bert} for notes on the English BERT pre-training update using whole-word masking.}

\begin{table*}[ht]
\centering
\begin{tabular}{@{}lccllcc@{}}
\toprule
\textbf{Model} & \textbf{Avg ; Modal ; Max} & \textbf{Target Noun}\\ & \textbf{Token Differences} & \textbf{\# Tokens}\\
\midrule
BETO & $\sim$ 0.682 ; 1 ; 3  & $\sim$ 1.07\\
BETO-uncased & $\sim$ 0.670 ; 1 ; 3  & $\sim$ 1.03\\
mBERT & $\sim$ 0.997 ; 1 ; 4 & $\sim$ 1.27\\
\midrule
DistilBETO & $\sim$ 0.670 ; 1 ; 3 & $\sim$ 1.03\\
\midrule
ALBETO-tiny & $\sim$ 0.619 ; 1  ; 3 & $\sim$ 1.04\\
ALBETO-base & $\sim$ 0.619 ; 1  ; 3 & $\sim$ 1.04\\
ALBETO-large & $\sim$ 0.619 ; 1 ; 3 & $\sim$ 1.04\\
ALBETO-xlarge & $\sim$ 0.619 ; 1 ; 3  & $\sim$ 1.04\\
ALBETO-xxlarge & $\sim$ 0.619 ; 1 ; 3 & $\sim$ 1.04\\
\midrule
RoBERTa-BNE-base & $\sim$ 0.643 ; 0 ; 3 & $\sim$ 1.05\\
RoBERTa-BNE-large & $\sim$ 0.643 ; 0 ; 3 & $\sim$ 1.05\\
XLM-RoBERTa & $\sim$ 0.929 ; 1 ; 4 & $\sim$ 1.28\\

\bottomrule
\end{tabular}
\caption{Average, modal, and maximum token differences across sentence pairs per LM. Tokenization schemes for all Spanish monolingual LMs were heavily represented by either non-zero or single-token differences across sentence pair stimuli, whereas the multilingual models tested here tended to more frequently generate non-zero token differences across the sentence pairs.}
\label{tab:lm_token_diffs}
\end{table*}

\subsection{Analysis of Expected Layer}\label{sec:appendix_expected_layer}

In the primary manuscript, we identified which layers of BETO provided the \textit{best} fit, i.e., which were most effective for predicting Sense Relationship (layer 6) and which were most effective at predicting Mean Relatedness judgments (layer 12). However, in some cases, the improvements across layers are fairly marginal. Thus, in this supplementary analysis, we implemented a version of the \textit{Expected Layer} analysis described by \citet{tenney-etal-2019-bert}. This analysis considers the size of the improvement across layers and estimates the layer at which particular kinds of information is expected to resolve in the network.

\subsubsection{Methods}

The \textit{Expected Layer} statistic considers the \textit{improvement in performance} (measured here as $AIC$ or $R^2$, depending on the analysis in question) across progressively more complex regression models fit with cosine distance information from each layer. This improvement in performance measure was defined as:

\begin{equation}\centering
    \Delta^\ell = Score(P_T^\ell) - Score(P_T^{\ell-1})
\end{equation}

Where $Score(P_T^\ell)$ is defined as the performance ($AIC$ or $R^2$) of a regression model equipped with cosine distance information from a given layer $\ell$ and each previous layer, i.e., such that the number of parameters in the regression model was equal to $\ell$. (Note that this was distinct from the approach taken in the primary manuscript, in which distinct univariate regression models were fit for each layer.) The Expected Layer statistic itself was defined as follows:

\begin{equation}\centering
\bar{E}_{\Delta}[\ell] = \frac{\sum_{\ell=1}^{L} \ell \cdot \Delta^{(\ell)}_T}{\sum_{\ell=1}^{L} \Delta^{(\ell)}_T}
\end{equation}

\subsubsection{Results}

Using this approach, we obtained Expected Layer statistics for both predicting Sense Relationship ($3.6$) and for predicting Mean Relatedness ($3.4$). Note that in both cases, the Expected Layer was smaller than the layer at which optimal performance was achieved; this is consistent with the observation that past a certain point, additional LM layers resulted in only marginal gains in prediction.


\begin{figure}
    \centering
    \includegraphics[width=0.75\linewidth]{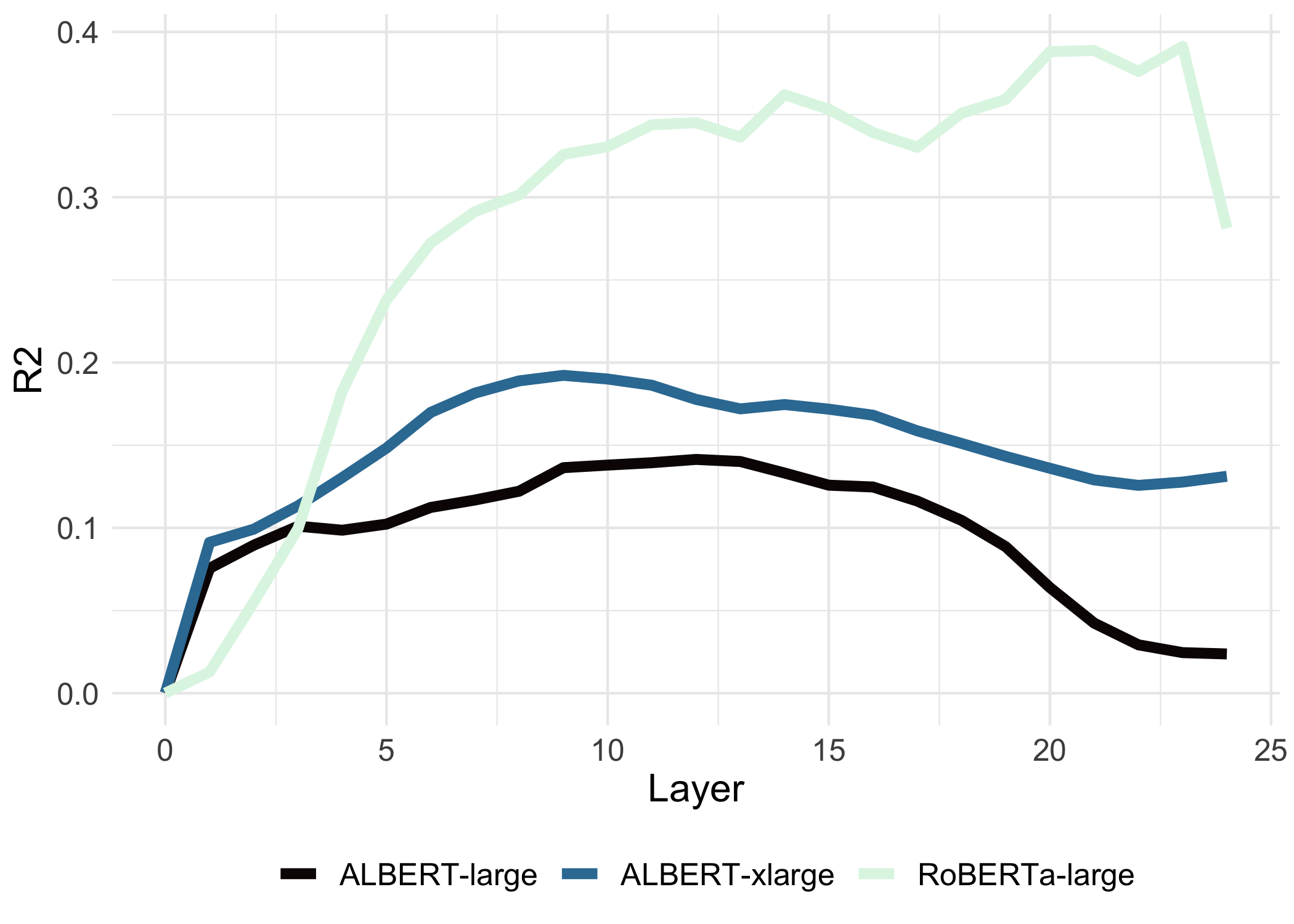}
    \caption{Depiction of three pre-trained Spanish LMs' ability to predict human relatedness judgments across layers (measured as $R^2$). For ease of illustration, this plot shows only LMs with $24$ layers.}\label{fig_appepndix:r2_by_layer}
\end{figure}

\subsection{Additional analyses of pre-trained English BERT-based models}

In the primary manuscript, we reported the results of work using ambiguity as a probe for understanding and interpreting how pre-trained Spanish LMs process word meanings. We found that larger models exhibited better performance; and secondly, that different model \textit{families} exhibited different trajectories of performance across layers. We asked whether these results replicated in pre-trained English models, using an openly available dataset of relatedness judgments about ambiguous English words, in context (RAW-C) \cite{trott-bergen-2021-raw}.

\subsubsection{A correlation between model scale and performance}

\begin{figure}
    \centering
    \includegraphics[width=0.85\linewidth]{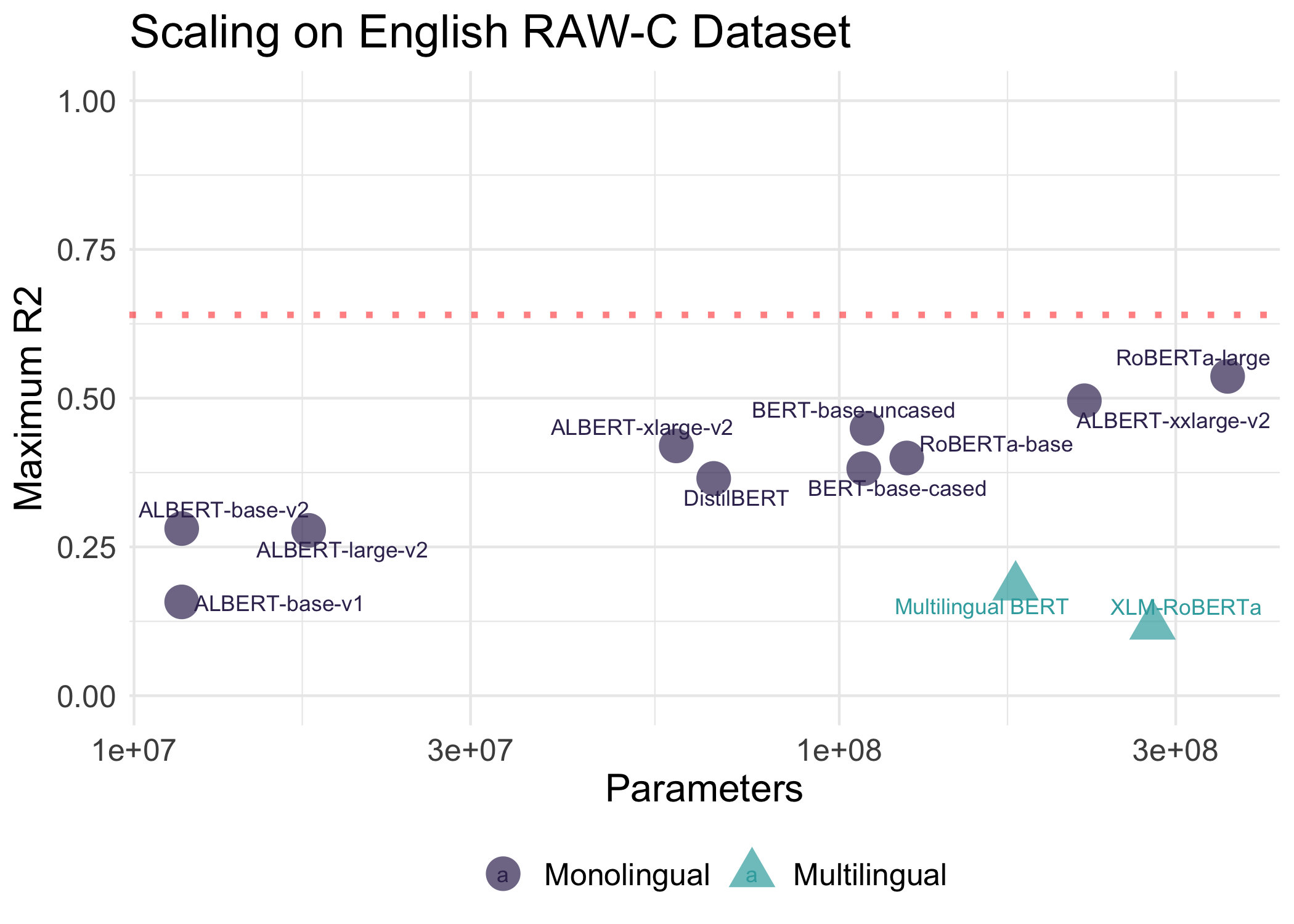}
    \caption{Pre-trained monolingual English language models show evidence of \textit{scaling}, i.e., models with more parameters achieve a higher $R^2$ in predicting human relatedness judgments about ambiguous English words.}\label{fig_appepndix:r2_english_scaling}
\end{figure}

We tested English versions of the Spanish language models tested in the primary manuscript (with the exception of \textit{ALBERT-tiny}, and with the addition of two different versions of \textit{ALBERT-base}). We then asked whether there was a relationship between the number of parameters in each model and the maximum $R^2$ achieved in predicting human relatedness judgments about English words, in context. Unlike Spanish, we found a clear, positive relationship between the logarithm of the number of parameters and the maximum $R^2$ (see Figure \ref{fig_appepndix:r2_english_scaling}): a linear model estimating maximum $R^2$ from number of parameters and a model's multilingual status estimated that for every order of magnitude increase in a model's number of parameters, $R^2$ increased by approximately $0.2$ $[\beta = 0.198, SE = 0.03, p < .001]$. On average, multilingual models also performed worse (adjusting for number of parameters), though the small number of multilingual models tested makes it difficult to determine whether this is a reliable finding.

\subsubsection{Layer-wise trajectories by model family}

\begin{figure}
    \centering
    \includegraphics[width=0.85\linewidth]{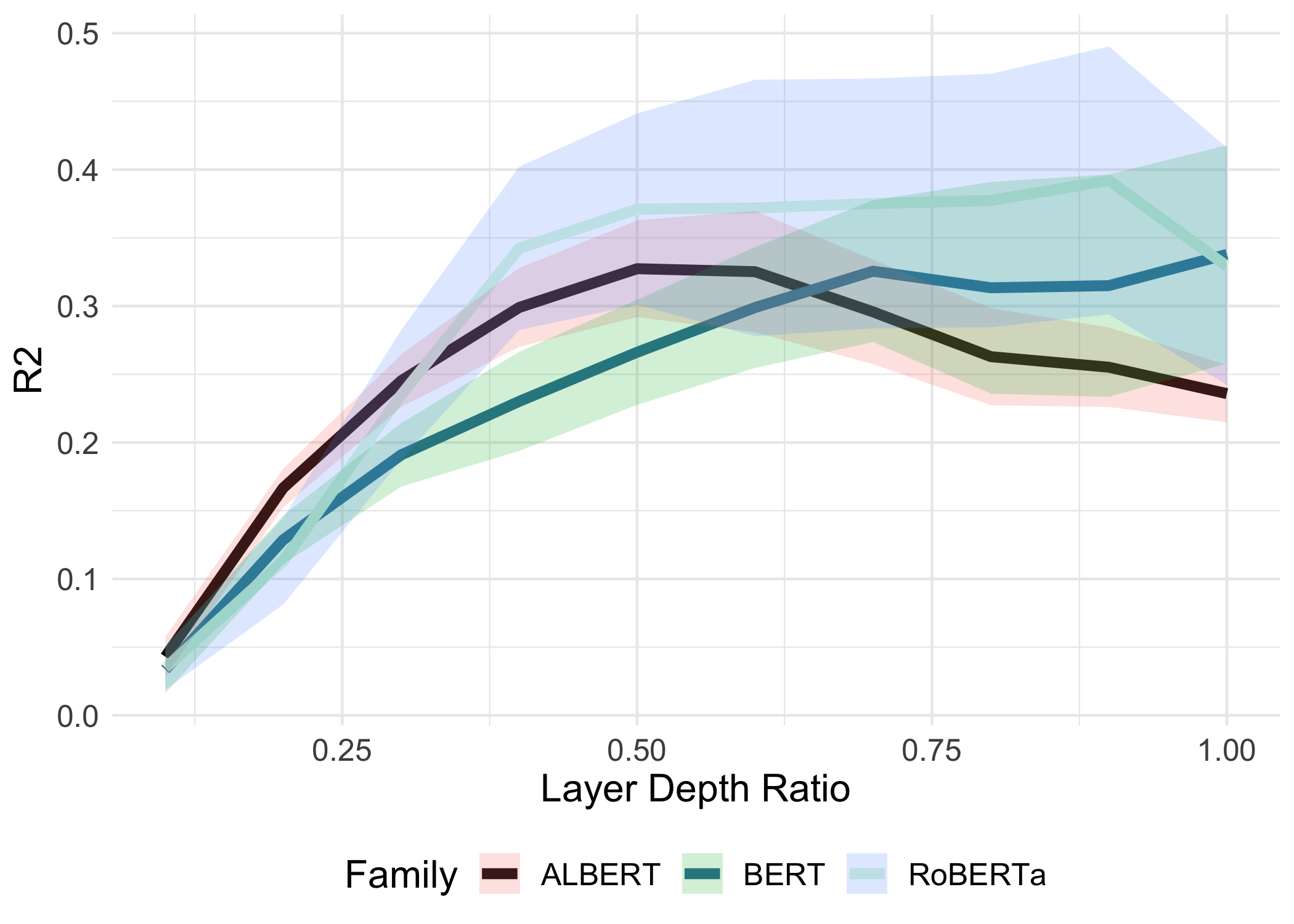}
    \caption{Depiction of English LMs' ability to predict mean relatedness judgments (measured as $R^2$), broken down by \textit{Model Family} and \textit{Layer Depth Ratio}, i.e., with each layer divided by the the total number of layers in a given model.}\label{fig_appepndix:r2_trajectories_english}
\end{figure}

We also asked whether different model \textit{families} displayed different performance trajectories across \textit{layers}. In the primary manuscript, we found that the ALBERT family models displayed a \textit{rise and fall} trajectory, while both BERT and RoBERTa displayed \textit{rise and plataeau} trajectories (see Figure \ref{fig:r2_by_prop_layer}). Surprisingly, we found qualitatively similar (albeit weaker) classes of trajectories using the pre-trained English models on the RAW-C dataset (see Figure \ref{fig_appepndix:r2_trajectories_english}).

\subsection{Analysis of GPT-4 Turbo}

In the primary manuscript, we found that the Large Language Models tested produced representations that were \textit{correlated} with human judgments, but nonetheless systematically underestimated how related people judged \textsc{Same Sense} meanings to be---and overestimated how related people judged \textsc{Different Sense} meanings to be (see Figure \ref{fig:sense_boundary_residuals}). However, recent work \cite{trott2024can, dillion2023can} suggests that state-of-the-art LLMs like GPT-4 are capable of producing ``norms'' that accurately predict human judgments across various domains, including the relatedness of English words. 

Because these models are ``closed source'', this work typically relies on \textit{prompting} the models with instructions and directly eliciting a judgment (e.g., a relatedness rating). Thus, a key limitation is that even if these judgments are highly correlated with human judgments, it is very difficult (in some cases impossible) to know \textit{why}, i.e., which representations or mechanisms give rise to the behavior in question---making them less well-suited to questions about model interpretability.

Nonetheless, the question of the empirical fit between these LM judgments and human judgments is still an interesting one---particularly because past work has primarily focused on judgments in English, and it is unclear whether these LMs would excel at other languages. Thus, in this supplementary analysis, we asked whether GPT-4 Turbo, a state-of-the-art LM, produced judgments that were more predictive of human judgments than the models tested in the primary manuscript. 

\subsubsection{Methods}

Following past work \cite{martinez2024using,trott2024can, dillion2023can}, we prompted GPT-4 Turbo (\textit{gpt-4-1106-preview}) using the OpenAI Python API. GPT-4 Turbo was presented with a system prompt containing the same instructions (in Spanish) that were presented to human participants, explaining the purpose of the task. Then, for each sentence pair, Turbo was presented with the same instructions given to human participants (again in Spanish) asking them to rate the relatedness of the target word across the two contexts. The two sentences were presented on separate lines, as was the target word (e.g., ``Word: aceite''). Finally, we included an additional instruction requesting a single number in response. Turbo was prompted using a temperature of 0 and its responses were limited to a maximum of 3 tokens.

\subsubsection{Results}

The ratings produced by Turbo were highly correlated with human judgments, approaching or even exceeding average human inter-annotator agreement ($\rho = 0.79$). 

We then asked whether Turbo's ratings explained away the sense boundary effect observed in the primary manuscript. First, we fit a linear model with Mean Relatedness as a dependent variable and two predictors: Rating (from GPT-4 Turbo) and Sense Relationship. The coefficients assigned to each predictor were significant, suggesting that they explained some amount of independent variance: both Rating $[\beta = 0.45, SE = 0.02, p < .001]$ and \textsc{Same Sense} $[\beta = 1.12, SE = 0.05, p < .001]$ exhibited a positive relationship. 

\begin{figure}
    \centering
    \includegraphics[width=0.75\linewidth]{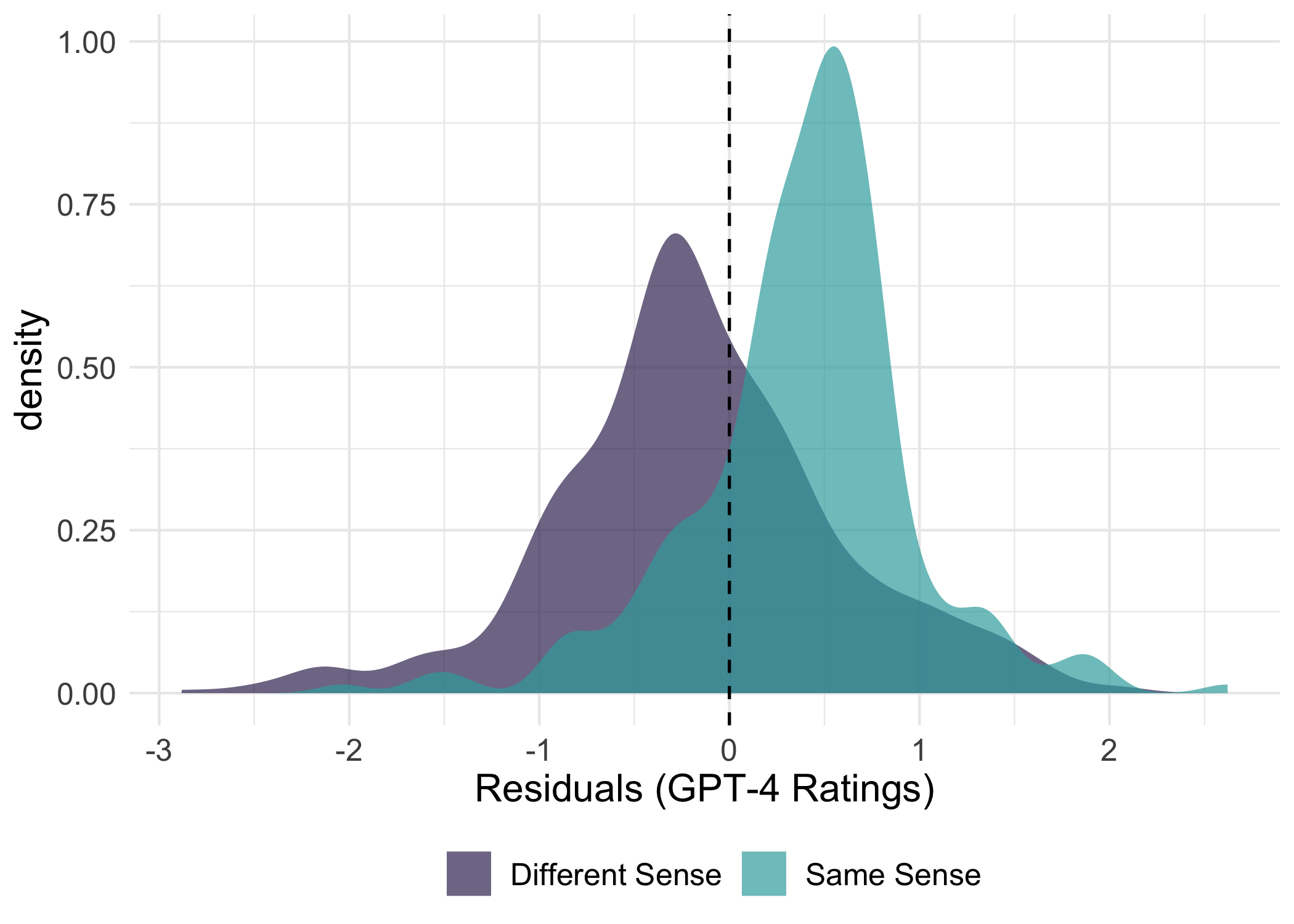}
    \caption{Residuals from a linear regression predicting Mean Relatedness from ratings elicited from GPT-4 Turbo. Although Turbo's ratings are highly correlated with human judgments, they still systematically underestimate the relatedness of Same Sense pairs and overestimate the relatedness of Different Sense pairs.}\label{fig_appepndix:gpt4}
\end{figure}

As in Figure \ref{fig:sense_boundary_residuals}, we also visualized the residuals of a linear model containing only Rating as a predictor (Figure \ref{fig_appepndix:gpt4}). Notably, even though Rating was highly correlated with Mean Relatedness, the residuals suggest that Turbo's ratings follow a similar pattern with respect to sense boundaries as was observed with BETO and the other models tested: GPT-4 consistently \textit{underestimates} the relatedness of \textsc{Same Sense} pairs, and consistently \textit{overestimates} the relatedness of \textsc{Different Sense} pairs.

\subsubsection{Discussion}

In this supplemental analysis, we found that a larger model trained with Reinforcement Learning from Human Feedback (RLHF) produced behavior that was much more correlated with human relatedness judgments than the other Spanish language models tested. 

On the one hand, this analysis has considerable limitations: as noted above, a major challenge with relying on large language models such as GPT-4 is that, despite their impressive performance, much is still unclear about how exactly they were trained---either in terms of the original training data or the procedure for implementing RLHF. Thus, as a tool for testing scientific hypotheses, they may be less useful than open-source models, i.e., the ones tested in the primary manuscript, especially if the scientific question under investigation concerns the \textit{representations} or \textit{mechanisms} used by the LLM. These are, of course, the kinds of questions that research on interpretability is generally interested in.

On the other hand, the fact that an LLM produces behavior that rivals inter-annotator agreement suggests that the representations required to produce this behavior can be learned provided sufficient training data and fine-tuning; in this case, direct data contamination is an unlikely concern given that the materials were entirely novel and the ratings had never been published before. This analysis is also notable in that it reveals that even a very large-scale model trained with RLHF appears to be less sensitive to sense boundaries than human judgments, as depicted in Figure \ref{fig_appepndix:gpt4}. 

However, because Turbo is a closed-source model, it is difficult to draw firm conclusions from these results precisely because we cannot investigate where or how this behavior emerges. For example, we cannot investigate which \textit{layer} of GPT-4 Turbo appears to be most helpful for producing high-quality relatedness judgments. Future work would thus benefit from investigating which architectural features, training objectives, or fine-tuning procedures are likely candidates for producing this improvement in performance, ideally in open-source Spanish language models.

\subsection{Additional sentence stimuli examples}\label{sec:additional-stims}

All sentence stimuli will be made available following publication. Here, we list additional examples from the dataset. Since target ambiguous nouns in Spanish are not always ambiguous in English, we mark the target noun's English translation in bolded italics within each parenthetical. 

\begin{description}[noitemsep]
    \item \textbf{1a.} Era un \textbf{banco} financiero. (\textit{It was a financial \textbf{bank}}.)
    \item \textbf{1b.} Era un \textbf{banco} exitoso. (\textit{It was a successful \textbf{bank}}.)
    \item \textbf{2a.} Era un \textbf{banco} c\'omodo. (\textit{It was a comfortable \textbf{bench}}.)
    \item \textbf{2b.} Era un \textbf{banco} de pl\'astico. (\textit{It was a plastic \textbf{bench}}.)
\end{description}

\begin{description}[noitemsep]
    \item \textbf{1a.} Disfrut\'o el \textbf{ba\~no} caliente. (\textit{[S/he] enjoyed the warm \textbf{bath}}.)
    \item \textbf{1b.} Disfrut\'o el \textbf{ba\~no} fr\'io. (\textit{[S/he] enjoyed the cold \textbf{bath}}.)
    \item \textbf{2a.} Disfrut\'o el \textbf{ba\~no} remodelado. (\textit{[S/he] enjoyed the remodeled \textbf{bathroom}}.)
    \item \textbf{2b.} Disfrut\'o el \textbf{ba\~no} privado. (\textit{[S/he] enjoyed the private \textbf{bathroom}}.)
\end{description}

\begin{description}[noitemsep]
    \item \textbf{1a.} Ten\'ia una \textbf{mu\~neca} fracturada. (\textit{[S/he] had a fractured \textbf{wrist}}.)
    \item \textbf{1b.} Ten\'ia una \textbf{mu\~neca} lesionada. (\textit{[S/he] had an injured \textbf{wrist} }.)
    \item \textbf{2a.} Ten\'ia una \textbf{mu\~neca} preciosa. (\textit{[S/he] had a lovely \textbf{doll}}.)
    \item \textbf{2b.} Ten\'ia una \textbf{mu\~neca} pl\'astica. (\textit{[S/he] had a plastic \textbf{doll}}.)
\end{description}

\end{document}